\documentclass{INTERSPEECH2023}
\usepackage{algpseudocode}
\usepackage{multirow}
\usepackage{diagbox}
\usepackage{cite}
\usepackage{algorithm}
\usepackage{CJKutf8}
\usepackage{url}
\usepackage{CJK}
\usepackage{float}
\usepackage{arydshln}
\usepackage{amsmath}
\usepackage{subfigure}
\usepackage{graphicx}

\interspeechcameraready

\title{Enhancing the Unified Streaming and Non-streaming Model with \\ Contrastive Learning}

\name{Yuting Yang, Yuke Li$^*$\thanks{$^*$ Corresponding author}, Binbin Du}
\address{NetEase Yidun AI Lab, Hangzhou, China}
\email{\small {$\left\{yangyuting04, liyuke, dubinbin\right\}$@corp.netease.com}}

\begin{document}

\maketitle
\begin{abstract}
The unified streaming and non-streaming speech recognition model has achieved great success due to its comprehensive capabilities. In this paper, we propose to improve the accuracy of the unified model by bridging the inherent representation gap between the streaming and non-streaming modes with a contrastive objective. Specifically, the top-layer hidden representation at the same frame of the streaming and non-streaming modes are regarded as a positive pair, encouraging the representation of the streaming mode close to its non-streaming counterpart. The multiple negative samples are randomly selected from the rest frames of the same sample under the non-streaming mode. Experimental results demonstrate that the proposed method achieves consistent improvements toward the unified model in both streaming and non-streaming modes. Our method achieves CER of 4.66\% in the streaming mode and CER of 4.31\% in the non-streaming mode, which sets a new state-of-the-art on the AISHELL-1 benchmark.
\end{abstract}
\noindent\textbf{Index Terms}: unified model, streaming speech recognition, contrastive learning

\section{Introduction}
Recently, the rapid advancements in deep neural networks have made end-to-end (E2E) automatic speech recognition (ASR) systems a mainstream focus of ASR research. Compared with the traditional speech recognition system, end-to-end ASR systems have the advantages of simpler system composition, training process, and decoding procedure. Various E2E ASR models have been explored in the literature, which can be roughly categorized into three main approaches: Connectionist Temporal Classification (CTC) models \cite{graves2006connectionist}, Transducer models \cite{battenberg2017exploring,graves2012sequence}, and attention-based encoder-decoder (AED) systems \cite{chan2016listen,vaswani2017attention,dong2018speech}. 

Streaming scenarios pose a challenge to state-of-the-art E2E ASR systems like Transformer \cite{vaswani2017attention} and Conformer \cite{gulati2020conformer}. The self-attention mechanism and the source-target multi-head attention require the full context of the sequences, resulting in significant latency, which in turn limits their usability in streaming scenarios. Previously, many efforts have been made to overcome these limitations. Chunk-based \cite{dong2019self,tsunoo2019towards} and look-ahead \cite{moritz2020streaming} methods are proposed to control the latency of the encoder module. Triggered attention (TA) \cite{moritz2019triggered, moritz2020streaming} and monotonic chunkwise attention (MoCHA) \cite{chiu2017monotonic, tsunoo2019towards} are used to restrict the context of the attention mechanism. In addition, two-pass models have been developed to improve the final performance by integrating supplementary modules that reevaluate the hypotheses generated by the first-pass streaming model \cite{sainath2019two,hu2020deliberation}.


The unified models \cite{zhang2020unified,wu2021u2++,gao2020universal,narayanan2021cascaded,liang2022fast,yu2020dual} provide a simple and efficient solution to the ASR tasks by learning and optimizing a single neural network for both streaming and non-streaming scenarios. During the inference process, the capabilities for different scenarios are enabled by configuring different decoding settings. In addition, previous literature \cite{yu2020dual} demonstrated that the performance of streaming mode benefits from the unified way with weight sharing and joint training. During the training process, knowledge distillation is employed to make the performance of the streaming mode benefit from the knowledge derived from the non-streaming mode \cite{liang2022fast,yu2020dual}. 
However, knowledge distillation improves the performance of streaming mode while the performance improvement of non-streaming mode is negligible. Enhancing the unified model's performance even further remains a challenge. 

In this work, we focus on improving the performance bottleneck of unified ASR models from the neural representation perspective. We expect the result of the streaming mode to be similar to that in the non-streaming mode. From the aspect of neural representation, when the encoded representation of streaming mode is close to its non-streaming counterpart, the content of their speech recognition results should be similar. Inspired by the recent progress of contrastive learning frameworks in computer vision \cite{chen2020simple,he2020momentum} and speech processing \cite{baevski2020wav2vec,schneider2019wav2vec}, we designed a simple yet effective method for unified ASR models to learn the representations that meet the above-mentioned condition explicitly. On the one hand, the contrastive task bridges the gap between the representations in streaming and non-steaming modes. On the other hand, the unified model benefits from uniform distribution, which is a significant property in contrastive learning \cite{wang2021understanding,wang2020understanding}.
Our contributions are as follows.
\begin{itemize}
    \item We develop a novel contrastive learning method to improve the performance of the unified streaming and non-streaming model, built upon the joint training framework.
    \item Our experiments on the AISHELL-1 benchmark demonstrated that the proposed method achieves 4.66\% in the streaming mode and 4.31\% in the non-streaming mode, outperforming the previous state-of-the-art models. 
    \item We show that the proposed method indeed learns unified representations for two modes and performs better to improve the accuracy of the unified model.
\end{itemize}

\section{Related work}
\subsection{Unifying streaming and non-streaming ASR}
Recently, there has been growing interest with the unified model since it not only reduces the resource consumption of model training and deployment but also improves the performance of streaming speech recognition. There are several unified models based on the Transducer network \cite{yu2020dual,narayanan2021cascaded,tripathi2020transformer}. For instance, dual-mode \cite{yu2020dual} developed a unified framework with shared weights and joint training mechanisms, and utilized knowledge distillation to further strengthen the performance of the streaming mode. Cascaded model \cite{narayanan2021cascaded} built a single E2E ASR model consisting of streaming and non-streaming encoders. The decoder module generates outputs by utilizing the output produced by either the streaming or non-streaming encoder. 

In addition, some solutions adopt attention-based approaches, such as Universal ASR \cite{gao2020universal}, U2 \cite{zhang2020unified}, and U2++ \cite{wu2021u2++} models. Both U2 \cite{zhang2020unified} and U2++ \cite{wu2021u2++} models utilized the hybrid CTC/Attention architecture and designed a dynamic chunks strategy to meet diverse latency requirements during inference. For decoding, U2 \cite{zhang2020unified} proposed a two-pass decoding method that involves a CTC-Decoder to generate initial hypotheses followed by an Attention-Decoder to finalizes the results with full context attention at the end of the input. Additionally, U2++ \cite{wu2021u2++} further enhanced the model with a bidirectional attention decoder and a spectral augmentation method named SpecSub. 

\subsection{Contrastive learning}
Contrastive learning is proposed to learn general data representations from unlabeled data, which aims at maximizing the similarity between anchors and their positive examples while keeping randomly sampled negative examples far away from it in the embedding space. Contrastive learning is extensively used in unsupervised or self-supervised domains, especially in the field of computer vision \cite{chen2020simple,he2020momentum}. Since no labels are available, negative pairs are formed by the anchor and randomly chosen samples from the minibatch, and the positive pair often consists of the author and its variants via data augmentation operators such as crop, color distortion, and Gaussian blur. Contrastive learning is widely used in the natural language processing tasks, such as sentence representation learning \cite{kong2019mutual} and machine translation \cite{wei2020learning}. Contrastive learning is also applicable to the area of speech processing and achieves promising results. For instance, wav2vec \cite{schneider2019wav2vec} learns general representations from raw audio via a contrastive task that requires distinguishing a true future audio sample from negatives, and then the representations serve as an input to a speech recognition system. In addition, wav2vec 2.0 \cite{baevski2020wav2vec} is a self-supervised framework that learns to identify the correct masked-out quantized representation with a contextualized representation from a Transformer model. 

Inspired by the contrastive learning framework, we propose to introduce a contrastive task into the unified model, where the positive pairs consist of the representations computed on chunk-based context (streaming) mode and full context (non-streaming) mode, encouraging the neural representation pairs generated in two modes near together.

\section{Method}
\subsection{Model architecture}
\begin{figure}[t]
  \centering
  \includegraphics[width=\linewidth]{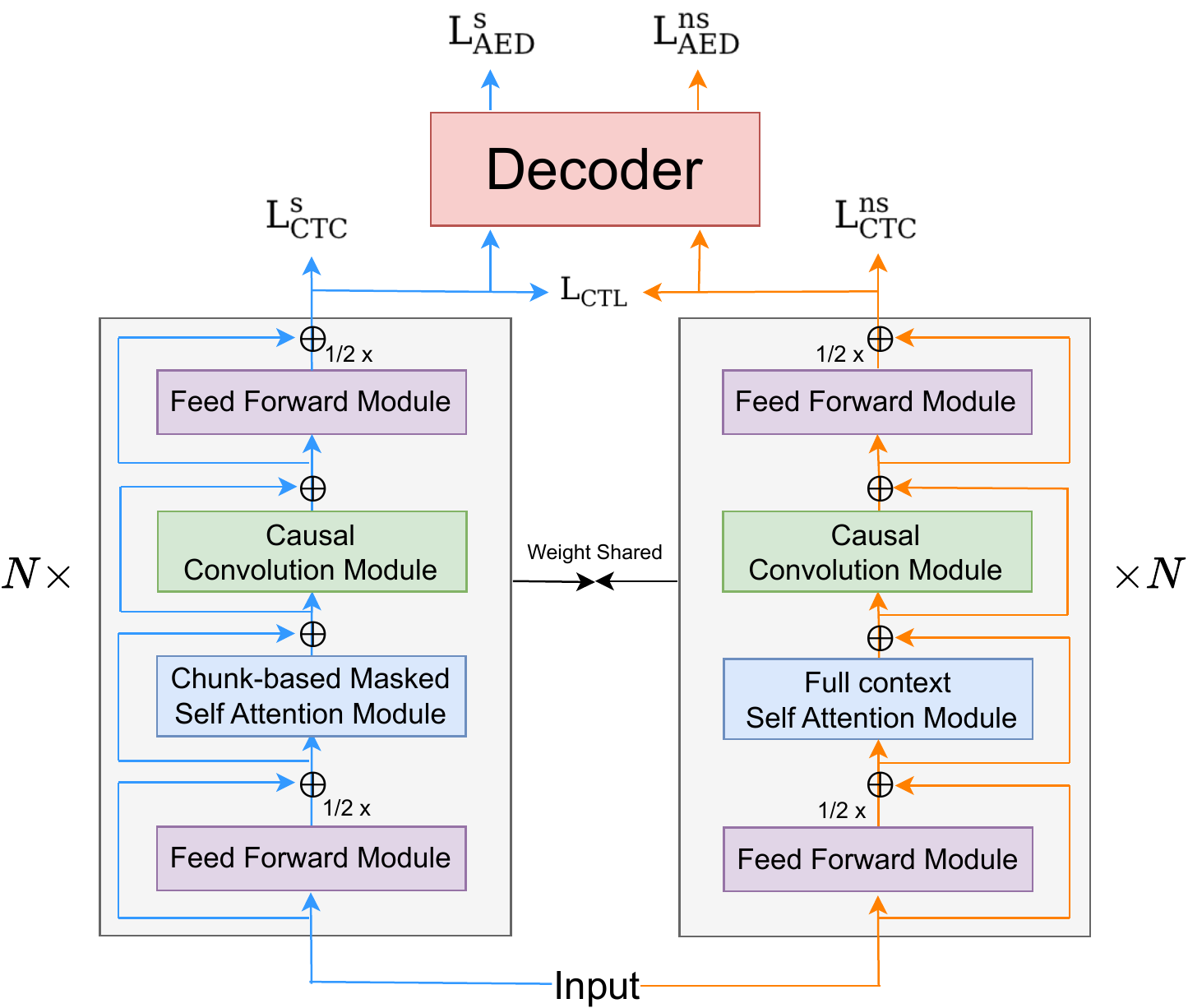}
  \caption{The framework of our method. $(\pounds_{CTC}^{s}, \pounds_{AED}^{s})$ denote the CTC loss and AED loss from the streaming branch (blue line). By contrast, $(\pounds_{CTC}^{ns}, \pounds_{AED}^{ns})$ denote the CTC loss and AED loss from the non-streaming branch (yellow line). $\pounds_{CTL}$ is the contrastive loss term. Layer normalization, the front convolution module, and the token embedding module are omitted. }
  \label{fig:framework}
  \vspace{-0.4cm} 
\end{figure}

The model architecture used in this paper is shown in
Figure \ref{fig:framework}. We adopt the same hybrid CTC/Attention \cite{watanabe2017hybrid} architecture as U2 \cite{zhang2020unified} and U2++ \cite{wu2021u2++} models. It mainly contains shared encoder and decoder modules. The former learns the context of the acoustic features in both full and chunk-based manners, and the latter predicts the next token given the high-level acoustic representations and previous token embeddings. We adopt the Conformer \cite{gulati2020conformer} encoder and the Transformer \cite{vaswani2017attention} decoder. An encoder block of the Conformer is composed of four modules, i.e., a feed-forward network, a self-attention module, a convolution module, and a second feed-forward network. In addition, the decoder layer of the Transformer is composed of three modules, a self-attention module, a multi-head attention module, and a feed-forward module. During training, the self-attention operation is restricted to the chunk-based feature context in the streaming mode via a predefined mask matrix. In addition, causal convolution \cite{oord2016wavenet} layers are used where the convolution window is shifted to the left to avoid additional latency. 

\subsection{Joint training}
we adopt a joint training strategy \cite{yu2020dual} for optimization. Specifically, the total loss function is a combination of the streaming ASR loss, the non-streaming ASR loss, and the contrastive loss. Formally,
\begin{equation}
  {\rm \pounds = \pounds_{ASR}^{s} + \pounds_{ASR}^{ns} +  \pounds_{CTL}},
  \label{eq1}
\end{equation}
where

\begin{equation}
\begin{split}
  \rm \pounds_{ASR}^{s}= \lambda\pounds_{CTC}^{s} + (1-\lambda)\pounds_{AED}^{s},
  \\
  \rm \pounds_{ASR}^{ns}= \lambda\pounds_{CTC}^{ns} + (1-\lambda)\pounds_{AED}^{ns}.
  \label{eq2}
\end{split}
\end{equation}

The first two terms in Eq.\ref{eq1} are streaming ASR loss $\pounds_{ASR}^{s}$ and non-streaming ASR loss $\pounds_{ASR}^{ns}$ built from the streaming and non-streaming branches, respectively. Specifically, as shown in Eq.{\ref{eq2}, the $\pounds_{ASR}^{s}$ term is a combination of CTC loss $\pounds_{CTC}^{s}$ and AED loss $\pounds_{AED}^{s}$, and the $\pounds_{ASR}^{ns}$ term is a combination of CTC loss $\pounds_{CTC}^{ns}$ and AED loss $\pounds_{AED}^{ns}$ as well. For the streaming branch, we use dynamic chunk size for different batches, which is sampled from 1 to 25 \cite{zhang2020unified}. In addition, the last element introduces a contrastive loss term $\pounds_{CTL}$ (see Section 3.3 for details).

\subsection{Contrastive loss for unified ASR models}
We apply the encoder module to extract contextual features and product high-level hidden representations. We define that $h^{s}$ and $h^{ns}$ denote the output of the encoder module generated by the streaming and the non-streaming modes, respectively. Figure 2 illustrates the proposed method with contrastive learning. The main idea of our method is to introduce a loss that brings positive pairs consisting of streaming representations $h^{s}$ and corresponding non-streaming representations $h^{ns}$ near together while pushing irrelevant pairs (negative pairs) far apart from each other. Specifically, the frame-level contrastive loss for $(h_i^s, h_i^{ns})$ is defined as
\begin{equation}
    \mathcal{\pounds}_{fctl}(h_i^s, h_i^{ns}) = -\log \frac{\exp(sim(h_i^{s},h_i^{ns})/\tau)}{\sum_{k_j \in {A}}\exp(sim(h_i^{s},k_{j})/\tau)}.
\label{eq:fctl}
\end{equation}

Given a positive example of such a representation pair $(h_i^{s}, h_i^{ns})$, we randomly pick a set of $N$ representations from the rest frames of $h^{ns}$ as negative examples. In Eq.\ref{eq:fctl}, $A$ includes $ h_i^{ns}$ and $N$ distractors, and $\tau$ is the temperature hyper-parameter. We compute the cosine similarity $sim(a, b)=\frac{a^Tb}{{\Vert}a{\Vert} \cdot {\Vert}b{\Vert}}$ between the streaming representation and the non-streaming representation. The overall contrastive loss can be formulated as
\begin{equation}
    \mathcal{\pounds}_{CTL} = \frac{1}{n} \sum_{i=1}^{n} \mathcal{\pounds}_{fctl}(h_i^s, h_i^{ns}),
\label{eq:ctl}
\end{equation}
where $n$ is the number of speech frames.

\begin{figure}[t]
  \centering
  \includegraphics[width=\linewidth]{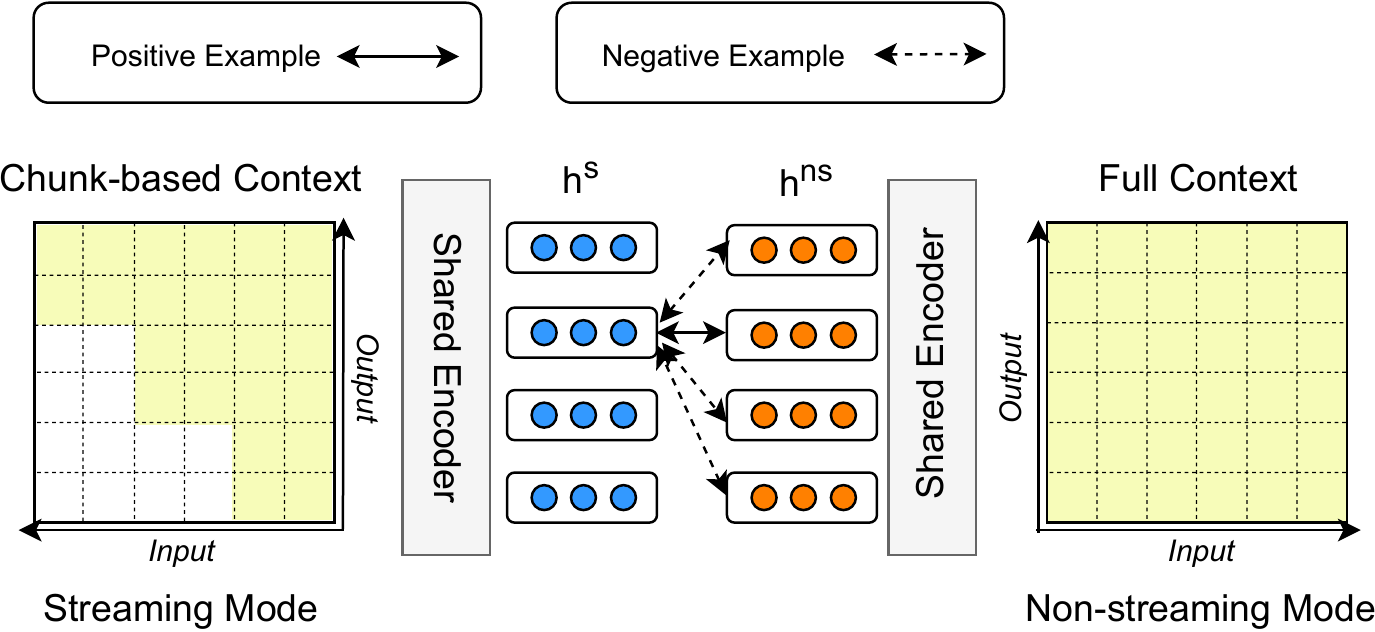}
  \caption{An illustration of the proposed method with contrastive learning.}
  \label{fig:speech_production}
\vspace{-0.4cm} 
\end{figure}

\section{Experiments}
\label{sec:experiments}

We conduct experiments on the AISHELL-1 \cite{bu2017aishell} corpus, containing over 170 hours of mandarin speech data from 400 speakers. The input speech features are 80-dimensional filterbank (FBank) features computed on 25ms windows with 10ms shifts. The vocabulary includes 4233 characters. 

\vspace{-0.1cm}
\subsection{Experimental setups}
We conduct experiments using the Wenet \cite{yao2021wenet} toolkit. We adopt the proposed method on two baseline models: U2 \cite{zhang2020unified} and U2++ \cite{wu2021u2++}. Two convolution downsampling layers with kernel size $3\times3$ and stride 2 in the front of the encoder are adopted. For U2 \cite{zhang2020unified}, it adopts a 12-layer Conformer encoder structure and a 6-layer Transformer decoder structure. The dimension of the attention layer is 256 with 4 split heads, and the dimension of the feed-forward layer is 2048. The kernel size is 15 for the Conformer encoder module. The U2 model has $\sim 46M$ parameters. For U2++ \cite{wu2021u2++}, The decoder module is a bidirectional structure with a 3-layer left-to-right decoder and a 3-layer right-to-left decoder. And the kernel size is 8 for Conformer. The other structure settings are the same as in U2. The U2++ model has $\sim 48M$ parameters.

\begin{table}[t]
  \caption{Character error rate (CER) on the AISHELL-1 test set. 1st pass and 2st pass stand for CTC prefix beam search and attention rescoring, respectively. $^\star$ is our reproduction. 
  }
  \label{tab:compare}
  \centering
  \tabcolsep=0.2cm
    \begin{tabular}{lccccc}
  \hline
    \multirow{2}*{Methods}&\multirow{2}*{Decoding mode}&\multicolumn{4}{c}{Decoding chunk size}\\
    \cline{3-6}
    &&Full&16&8&4 \\
	\hline
	\multirow{2}*{U2\cite{zhang2020unified}}&1st pass&5.51&6.23&6.57&6.92\\
    &2st pass&5.05&5.48&5.66&5.85\\
	\multirow{2}*{U2++\cite{wu2021u2++}}&1st pass&5.19&5.81&-&-\\
    &2st pass&4.63&5.05&-&-\\
    \multirow{2}*{U2++$^\star$}&1st pass&5.23&5.87&6.15&6.37\\
    &2st pass&4.66&5.05&5.23&5.39\\
	\hline
    \multicolumn{4}{l}{\textbf{Our method  with}} \\
	\multirow{2}*{{U2}}&1st pass&5.12&5.84&6.18&6.53\\
    &2st pass&4.76&5.21&5.43&5.63\\
	\multirow{2}*{{U2++}}&1st pass&5.03&5.65&5.87&6.16\\
    &2st pass& \bf 4.52& \bf 4.90& \bf 5.09& \bf 5.18\\
  \hline
  \end{tabular}
\vspace{-0.4cm}
\end{table}

We apply speed perturbation \cite{ko2015audio} and SpecAugment \cite{park2019specaugment} to the training data. U2++ based models additionally adopt the Specsub \cite{wu2021u2++} method. For training, we keep almost the same training settings as in the AISHELL-1 recipe, including regularization, optimizer, learning rate schedule, and data augmentation. Please refer to the open-sourced recipe \footnote{https://github.com/wenet-e2e/wenet/tree/main/examples/aishell/s0} for more details. We train the model for 240 epochs and 360 epochs on 8 V100 GPUs with U2 models and U2++ models respectively, and a final model is obtained by averaging the top 30 models with the best validation loss. The $\lambda$ in Eq.{\ref{eq2}} is set to 0.3 according to the recipe.
Additionally, we choose the temperature hyper-parameter in contrastive loss ranging from 0.05 to 0.8, and we set $\tau=0.4$ for the U2 model and $\tau=0.6$ for the U2++ model that achieve better performance than other options. Following \cite{baevski2020wav2vec}, we use $N=100$ distractors in the contrastive loss. For decoding, we present the first-pass (CTC prefix beam search \cite{hannun2014first}) results and the second-pass (attention rescoring \cite{zhang2020unified}) results, respectively. In Table~\ref{tab:online} and Table~\ref{tab:offline}, a 3-gram language model (LM) was trained on the transcripts used for shallow fusion.

\subsection{Results}
\subsubsection{Comparison with baseline models}
Table~\ref{tab:compare} compares the (CER) results of our method on the AISHELL-1 test set with baseline models \cite{zhang2020unified,wu2021u2++}. We can see from Table~\ref{tab:compare} that the proposed method outperforms the baseline models while using exactly the same architectures, demonstrating that contrastive learning gives a consistent improvement over the baseline model in both the streaming and non-streaming modes. The performance of the non-streaming mode is also improved by our framework with contrastive loss. Our explanation is that the contrastive loss makes the positive pair closer while pushing the irrelevant pairs farther. Therefore, the non-streaming mode benefits from the uniform distribution.

\vspace{-0.1cm}
\subsubsection{Comparison with published streaming models}
Table~{\ref{tab:online} shows the comparisons with recently published streaming systems and unified systems with streaming mode on the AISHELL-1 test set. Specifically, our method achieves CER of 4.9\% without external LM and CER of 4.66\% with a 3-gram LM for shallow fusion on the AISHELL-1 test set, surpassing recently published streaming models. To our best knowledge, this is the state-of-the-art performance for streaming ASR systems on the AISHELL-1 task.

\begin{table}[t]
  \caption{Comparison with recently published streaming E2E systems on the AISHELL-1 test set. $\Delta$ denotes the additional latency introduced by rescoring the first-pass hypotheses.
  }
  \label{tab:online}
  \centering
  \begin{tabular}{p{3.8cm}cc}
   \hline
    Methods &Latency (ms)  & Test set  \\
   \hline
    HS-DACS Transformer \cite{li2021head} &1280 &6.80 \\
    WNARS (w/ LM) \cite{wang2021wnars}&640&5.22 \\
    U2 \cite{zhang2020unified}&640 + $\Delta$&5.42\\
    U2++ \cite{wu2021u2++}&640 + $\Delta$&5.05\\
    CUSIDE (w/ LM) \cite{an2022cuside} &400 + 2&5.47 \\
    \hspace{0.2cm}+NNLM rescoring &400 + 2 + $\Delta$&4.79 \\
   \hline
    \textbf{Our method with} \\
    {U2} &640 + $\Delta$&5.21 \\ 
    \hspace{0.2cm}+ w/ a 3-gram LM&640 + $\Delta$&{4.87} \\
    {U2++} &640 + $\Delta$&4.90 \\ 
    \hspace{0.2cm}+ w/ a 3-gram LM&640 + $\Delta$&\textbf{4.66} \\
   \hline
  \end{tabular}
\end{table}

\begin{table}[t]
  \caption{Comparison of our unified model with pure non-streaming E2E systems on AISHELL-1. $^\star$ denotes our reproduction using the Conformer backbone with 46M parameters. }
  \label{tab:offline}
  \centering
  \begin{tabular}{p{3.8cm}cc}
   \hline
    Models  & Dev set & Test set  \\
   \hline
    Pure non-streaming $^\star$ &4.29&4.67\\
    Pure non-streaming \cite{guo2021recent} &4.4&4.7  \\
    
   \hline
    \textbf{Our method with} \\
    {U2} &4.36&4.76 \\ 
    \hspace{0.2cm}+ w/ a 3-gram LM&4.14&4.51 \\
    {U2++} &4.23&4.52 \\ 
    \hspace{0.2cm}+ w/ a 3-gram LM&\textbf{4.08}&\textbf{4.31} \\
   \hline
  \end{tabular}
\vspace{-0.3cm}
\end{table}

\vspace{-0.1cm}
\subsubsection{Comparison with pure non-streaming models}
Table~{\ref{tab:offline}} further compares the accuracy of our unified model with pure non-streaming models on the AISHELL-1 dataset. Specifically, our unified method achieves CER of 4.23\%/4.52\% without external LM and CER of 4.08\%/4.31\% with a 3-gram LM for shallow fusion on the AISHELL-1 dev/test sets under the non-streaming mode. These results demonstrate that our unified models obtain comparable results to pure offline models while having the benefit of comprehensive capabilities on streaming and non-streaming scenarios.

\vspace{-0.1cm}
\subsection{Discussions}
\subsubsection{Is contrastive loss more effective?}
We attempt to introduce the contrastive loss term to close the distance between streaming and non-streaming representations, which are obtained by restricted context and full context of input sequences, respectively. However, there are other options to achieve this goal from the idea of knowledge distillation, such as L2 loss. It bridges the gap by directly reducing the Euclidean distance between the representations of streaming and non-streaming modes, and no negative samples were introduced in this process. 


\begin{figure}[htbp]
\centering
\subfigure[w/o contrastive loss]{
\begin{minipage}[t]{0.45\linewidth}
\centering
\includegraphics[height=3.5cm,width=3.5cm]{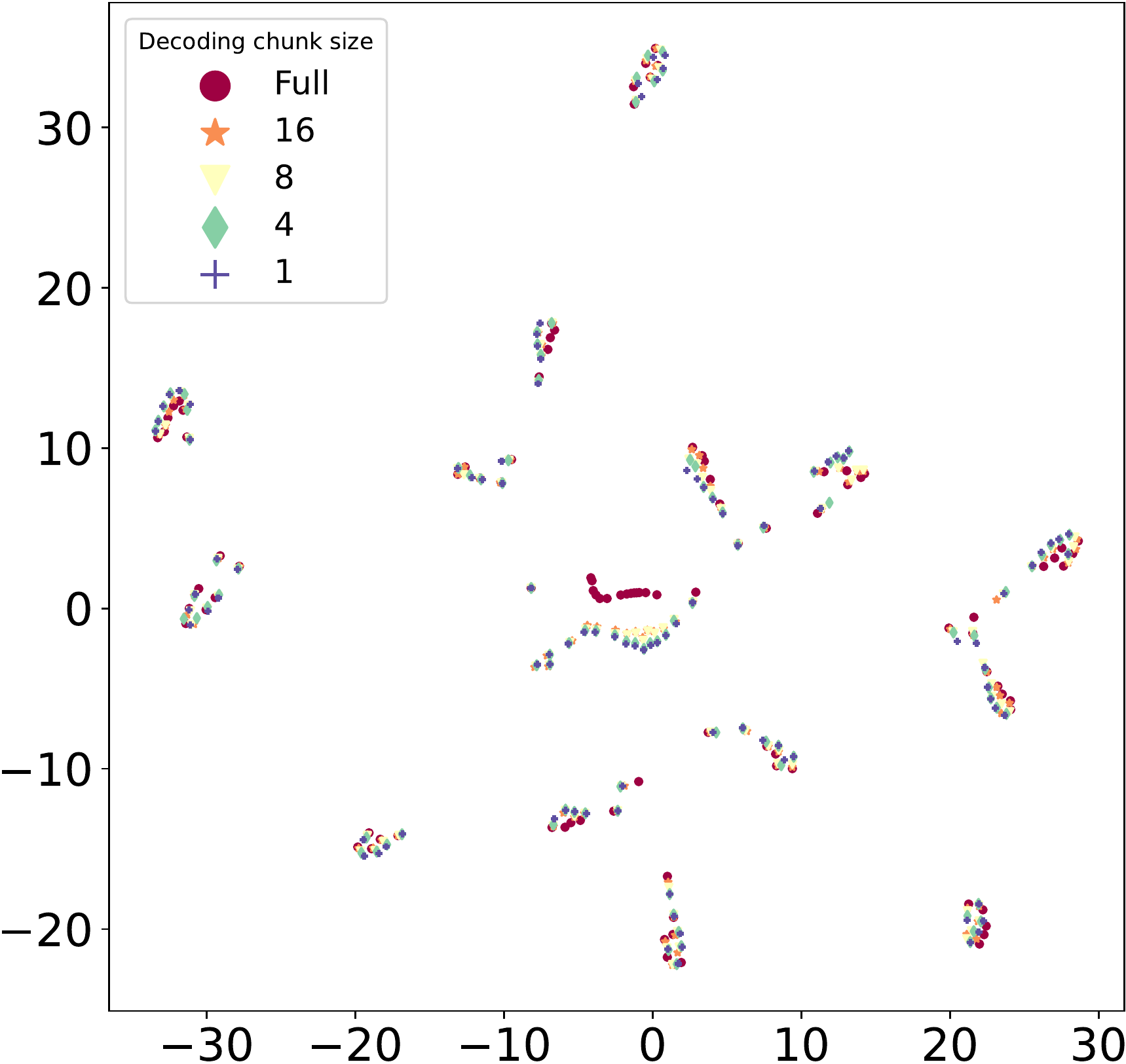}
\end{minipage}%
}%
\subfigure[w/ contrastive loss]{
\begin{minipage}[t]{0.45\linewidth}
\centering
\includegraphics[height=3.5cm,width=3.5cm]{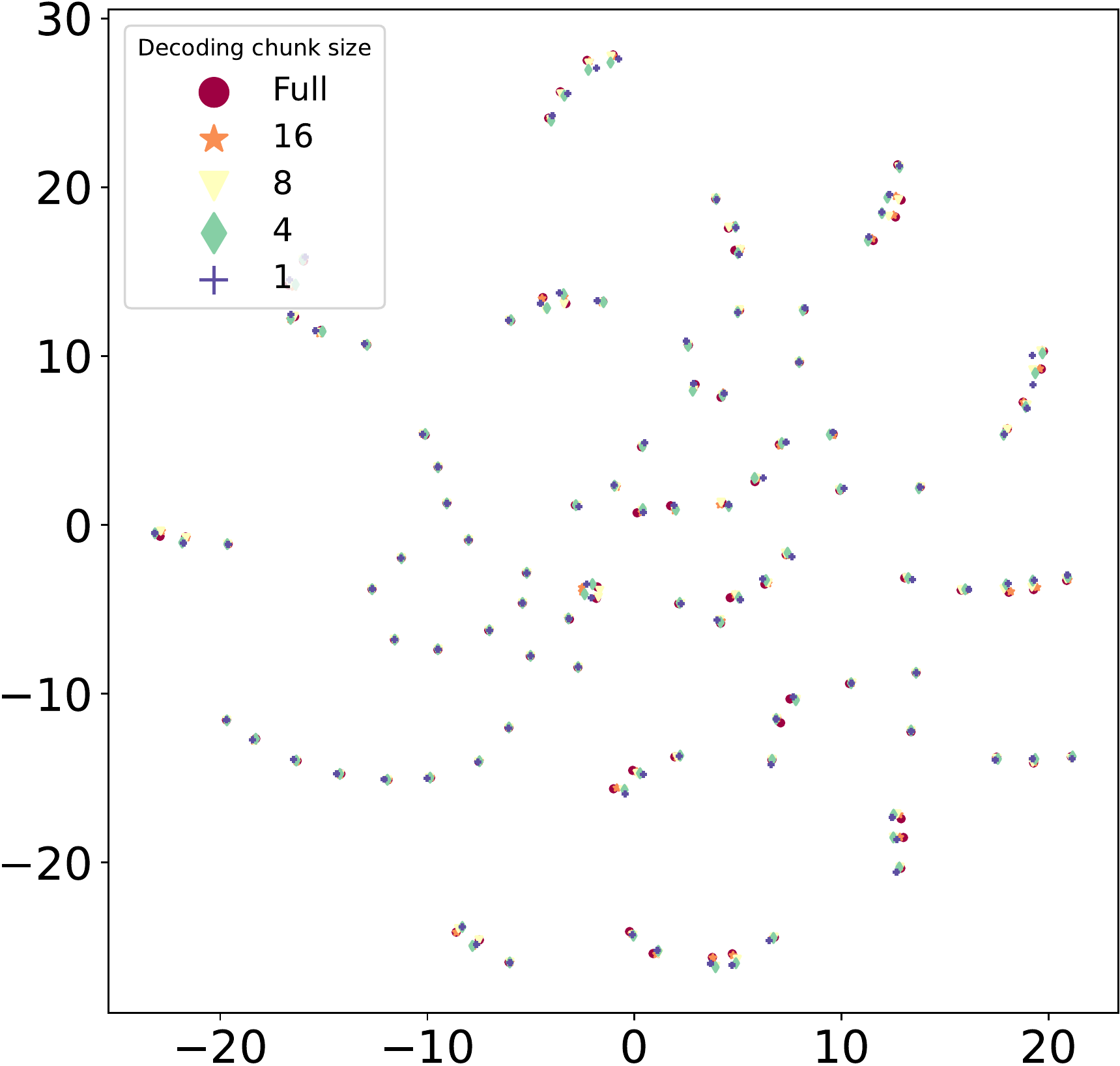}
\end{minipage}%
}%
\centering
\caption{Visualization of the hidden representation of streaming and non-streaming modes. We utilize T-SNE to reduce the dimension into 2D. The red dots are the encoded representations from non-streaming mode. In contrast, the other points stand for the representations from the streaming mode with decoding chunk size $\left\{16,8,4,1\right\}$, respectively. The utterance is randomly selected from the test set.} 
\label{fig:fig3}
\end{figure}

\begin{table}[t]
  \caption{Character error rate (CER) on the AISEHLL-1 test set under different loss terms settings with the U2 \cite{zhang2020unified} model. 1st pass and 2st pass decoding mode stand for CTC prefix beam search and attention rescoring, respecitively.
  }
  \label{tab:ablation}
  \centering
  \begin{tabular}{cccccc}
   \hline
    \multirow{2}*{Contrastive Loss} & \multirow{2}*{L2 Loss}&\multicolumn{2}{c}{1st pass}&\multicolumn{2}{c}{2st pass} \\
    && Full & 16&full&16  \\
   \hline
    Yes&No&\textbf{5.12}&\textbf{5.84}&\textbf{4.76}&\textbf{5.21} \\
     No&Yes&5.25&5.94&4.87&5.28 \\
    {No}&No&5.32&6.03&4.87&5.33 \\ 
    \hline
    
  \end{tabular}
\vspace{-0.3cm}
\end{table}

We study the effects of different loss terms based on the U2 model architecture. The results are shown in Table 4. Both the proposed method and L2 loss improve the accuracy of the unified model, and the proposed method performs better to this goal. These results in Table 4 suggest the effectiveness of incorporating the contrastive loss term in enhancing the performance of the unified model. 

\vspace{-0.1cm}
\subsubsection{Analysis on the gap between two modes}
Figure~\ref{fig:fig3} visualizes embedding spaces learned by the standard U2 model without contrastive loss and the proposed model with contrastive loss. Figure~\ref{fig:fig3}(a) shows that the representations are dissimilar. The gap exists between the two modes without any explicit terms to make them together. In contrast, as shown in Figure~\ref{fig:fig3}(b), the proposed model with contrastive learning is able to bring the representations of streaming and non-streaming modes closer. In addition, the representations are more uniformly distributed in the embedding space. These mean that the proposed method indeed encourages the representation pairs generated in two modes near together, thus improving the accuracy of the unified model.

\section{Conclusions}
In this paper, we propose to enhance the unified streaming and non-streaming ASR model with contrastive learning. Our method aims to bridge the representation gap between the full-context and chunk-based context modes. Major contributions are two-fold: (1) our method introduced a simple yet effective contrastive objective to make the representations at the same frame under two different modes to be similar; (2) extensive experiments demonstrate that our method set a new state-of-the-art on the widely-used AISHELL-1 benchmark. We demonstrated the effectiveness of the proposed method on the attention-based architecture, while our approach can be applied to other E2E ASR models, such as CTC-based and RNN-T models, which will be interesting future work. In addition, we will extend our method to more large-scale datasets in future work.

\clearpage
\bibliographystyle{IEEEtran}
\bibliography{mybib}

\end{document}